\documentclass[conference]{IEEEtran}

\usepackage{amsmath,amssymb,amsfonts}
\usepackage{algorithmic}
\usepackage{graphicx}
\usepackage{subcaption}
\usepackage{xcolor}

\begin{document}

\title{\textit{toon2real:} Translating Cartoon Images to Realistic Images\\
}

\author{
\IEEEauthorblockN{K. M. Arefeen Sultan\IEEEauthorrefmark{1},
Mohammad Imrul Jubair\IEEEauthorrefmark{2},\\
MD. Nahidul Islam\IEEEauthorrefmark{3},
Sayed Hossain Khan\IEEEauthorrefmark{4}
}
\IEEEauthorblockA{
Department of Computer Science and Engineering,\\
Ahsanullah University of Science and Technology,
Bangladesh\\
\{\IEEEauthorrefmark{1}krsultan069,
\IEEEauthorrefmark{3}nahidul19967,
\IEEEauthorrefmark{4}sayedhossainkhan36\}@gmail.com,
\IEEEauthorrefmark{2}mohammadimrul.jubair@ucalgary.ca
}
}

\maketitle

\begin{abstract}
In terms of Image-to-image translation, Generative Adversarial Networks (GANs) has achieved great success even when it is used in the unsupervised dataset. In this work, we aim to translate cartoon images to photo-realistic images using GAN.
We apply several state-of-the-art models to perform this task; however, they fail to perform good quality translations.
We observe that the shallow difference between these two domains causes this issue. Based on this idea, we propose a method based on CycleGAN model for image translation from cartoon domain to photo-realistic domain. To make our model efficient, we implemented Spectral Normalization which added stability in our model. We demonstrate our experimental results and show that our proposed model has achieved the \textit{lowest Fr\'echet Inception Distance score} and better results compared to another state-of-the-art technique, UNIT.
\end{abstract}


\section{Introduction}
\begin{figure}[!htb]
    \centering
    \begin{subfigure}[b]{0.2\textwidth}
        \includegraphics[scale=0.75]{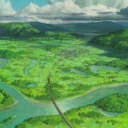}
        \caption{Cartoon scene}
        \label{subfig: intro1}
    \end{subfigure}
    \begin{subfigure}[b]{0.2\textwidth}
        \includegraphics[scale=0.75]{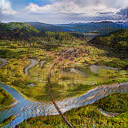}
        \caption{Our result}
        \label{subfig:intro2}
    \end{subfigure}
    \caption{An example of cartoon to real world translation. (a) \textit{Input image}: which is from the animated film "The Wind Rises". (b) \textit{Our result}: transforming the cartoon image (a) to real world image.}
    \label{fig:intro}
\end{figure}

Cartoons occupy a huge part in our entertainment sector.
Film industries, in recent days, are remaking movies from the popular past cartoons and presenting them for current generation. Such an example is --
the \textit{The Lion King (2019)} from \textit{The Lion king (1994)}. Therefore, we realize the necessity of recreating realistic images from the cartoons which can contribute to photo-realistic rendering in computer graphics as well as in film industries.
In this paper, we propose an approach which converts images from cartoons into their corresponding photo-realistic images. From Figure~\ref{fig:intro}, we can see an outcome of our work where the cartoon scene is translated into a photo-realistic one.

Image-to-image translation using Generative Adversarial Network (GAN)\cite{GoodfellowPMXWOCB14} has been one of the most desiring fields of deep learning research lately. In a GAN architecture, a discriminator network tries to measure the probability of whether an image has come from an authentic data source or a fake generated source of the generator. On the other hand, the generator tries to maximize the probability of the discriminator's making mistake and while doing that, it learns to generate more accurate data as much as close to real data. 
Tremendous success of GANs \cite{GoodfellowPMXWOCB14}, led other researchers to work on unsupervised settings of datasets, such as \cite{ZhuPIE17}, \cite{KimCKLK17}. Although these models have succeeded to translate one domain of image to another in general, there hasn't been any specific research for generating images of the photo-realistic domain from cartoon domain.
This is an extremely hard task; the reason is the domain gap between these two distributions is too shallow.

As a result, the discriminator can be easily erroneous to determine the generated data as real ones.
This is the reason why most state-of-the-art models tend to fail in case of generating cartoon to real images. We illustrate in our result section that several models intent to keep the original content of cartoon domain while generating photo-realistic images.

To satisfy our objective on this task, we have taken an approach built on CycleGAN\cite{ZhuPIE17}. We implemented spectral normalization technique\cite{abs-1802-05957} which helps our model to converge faster. Our approach also keeps the content of photo-realistic domain. In addition to these, we also created our own dataset for our model.
We show that our method has the lowest FID score than the other baseline models, and also it tends to show more stabilization in training than the others.

\section{Related Works}
In this section, we review on different relevant variations of GAN and several past works on image-to-image translations.

GANs have achieved great results in various image generation tasks, which are image super-resolution\cite{JohnsonAL16}, image-to-image translation\cite{ZhuPIE17,abs-1810-04991,LiuBK17}, text-to-image synthesis\cite{ZhangXLZHWM16,ReedAYLSL16} etc. 
Recently, GAN\cite{GoodfellowPMXWOCB14} based approach has given tremendous results in image-to-image translation tasks. \textit{Zhu et al.}\cite{ZhuPIE17} proposed a cycle consistency loss to reduce the infinite mappings of input images to any distribution in the target domain. \textit{Adversarial loss} alone can't solve the random permutation mappings of target distribution, rather it helps the input image to be translated into target domain. Similar to CycleGAN, \textit{Kim et al.} \cite{KimCKLK17} proposed a method for preserving the key attributes between the input and the transformed image while maintaining a cycle consistency criterion. Similarly, \textit{Yi et al.} \cite{yi2017dualgan} proposed dual-GAN mechanism based on dual learning from natural language translation\cite{he2016dual}.
In UNIT\cite{LiuBK17} framework, \textit{Liu et al.} proposed a shared-latent space assumption, which denotes that the pair of corresponding images in different domains can be mapped to a same latent representation in a shared-latent space. \textit{Liu et al.} used the combination of generative adversarial network (GAN), based on CoGAN\cite{liu2016coupled} and variational autoencoders (VAEs)\cite{kingma2013auto,larsen2015autoencoding,rezende2014stochastic}.\\
For stabilizing and improving the training of GAN, several works were proposed such as adding weight normalization and regularization techniques \cite{GulrajaniAADC17, abs-1802-05957}, designing new generative architectures\cite{RadfordMC15, karras2017progressive} to improve visual results and modifying learning objectives \cite{Arjovsky2017WassersteinG,metz2016unrolled}.
\textit{Miyato et al.}\cite{abs-1802-05957} first proposed spectral normalization technique which constrains the Lipschitz constant of the discriminator network by limiting the spectral norm of each layer. The authors argue that spectral normalization can improve the quality of training GANs better than weight normalization\cite{salimans2016weight} and gradient penalty\cite{GulrajaniAADC17}. In this paper, we have utilized this property of spectral norm to improve the training of our GAN.\\
In our previous work \textit{Cartoon-to-real} \cite{abs-1811-11796}, CycleGAN based model was implemented to transform cartoon images into photo-realistic domain. While all these methods achieve compelling results, they take too much time for training. The reason behind it is that these models are not stabilized during training. In the next section, we discuss the approach we took to solve these fundamental issues and to obtain better outcomes.

\section{Formulation}
Our main objective is to transform \textit{cartoon} images to \textit{photo-realistic} images by learning the mapping of a \textit{cartoon} domain $C$ to the \textit{photo-realistic} domain $R$. In this section we discuss about the approaces we have taken to accomplish the task.

\begin{figure}
\centering
    \begin{subfigure}[b]{0.5\textwidth}
        \includegraphics[width=\textwidth]{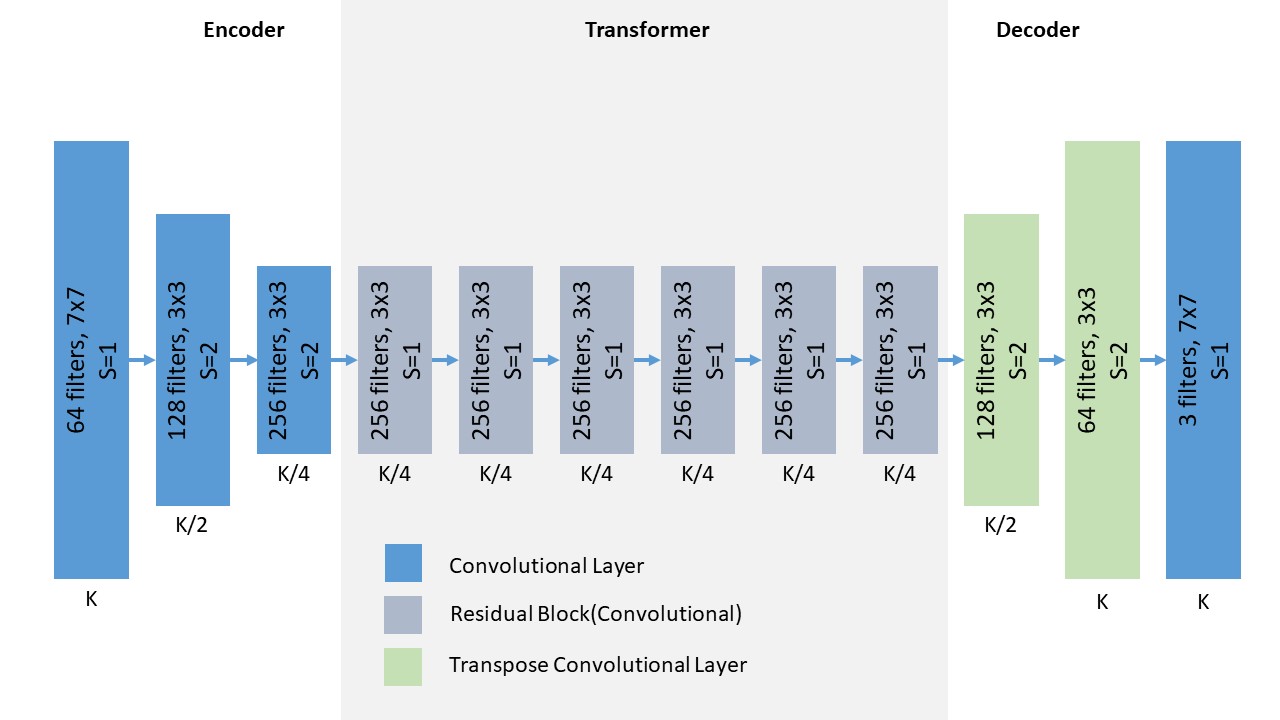}
        \caption{Generator Network}
    \end{subfigure}
    
    \begin{subfigure}[b]{0.5\textwidth}
        \includegraphics[width=\textwidth]{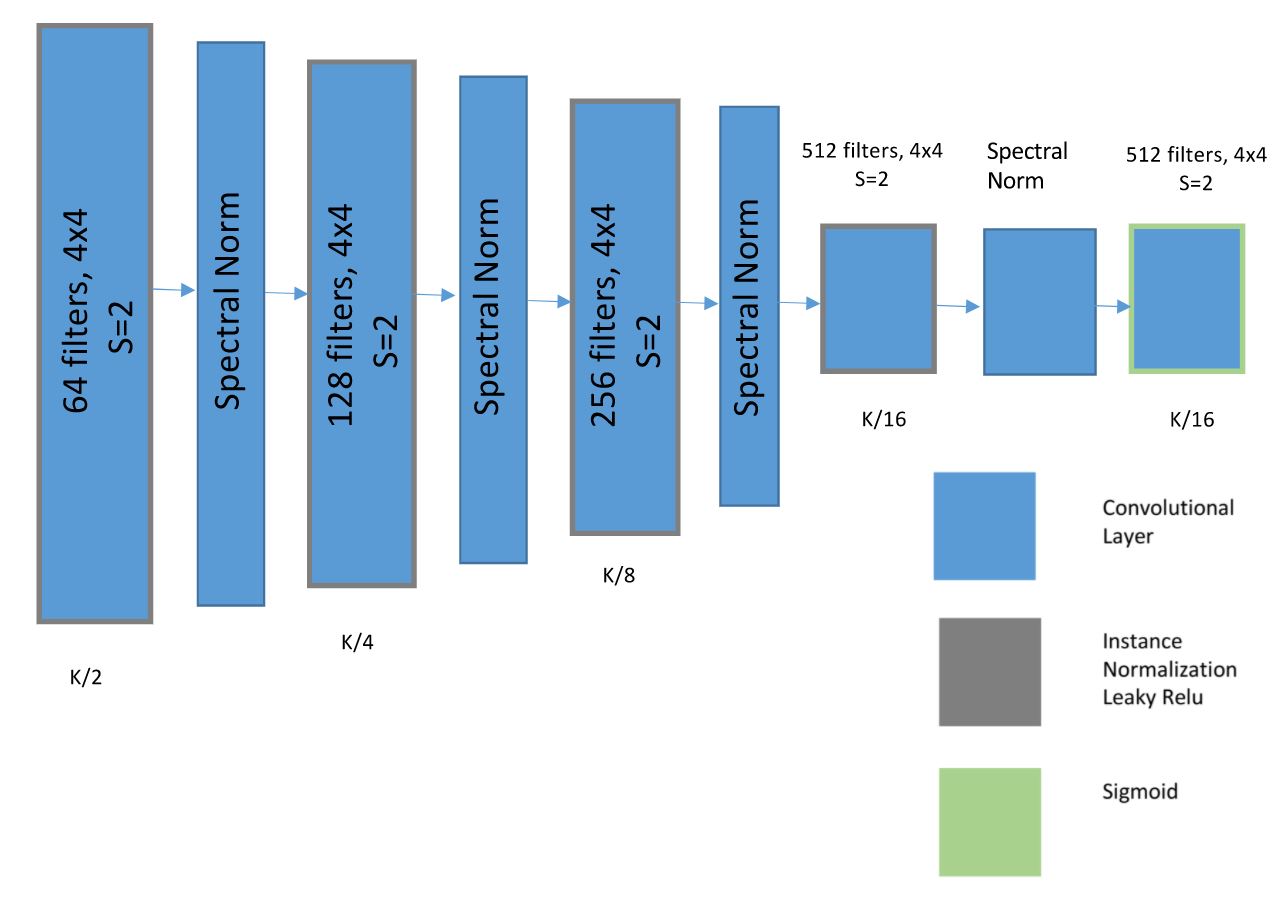}
        \caption{Discriminator Network}
    \end{subfigure}
\caption{Architectures of the generator and discriminator of our proposed model, in which \textit{k} is the kernel size and \textit{s} is the stride in each convolutional layer.}
\label{fig:cycle-gan-archi}
\end{figure}

\begin{figure*}[!htb]
    \centering
    \begin{subfigure}[b]{0.2\textwidth}
        \centering
        \includegraphics[scale=0.7]{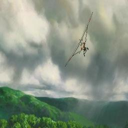}
    \end{subfigure}
    \begin{subfigure}[b]{0.2\textwidth}
        \centering
        \includegraphics[scale=0.7]{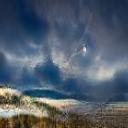}
    \end{subfigure}
    \begin{subfigure}[b]{0.2\textwidth}
        \centering
        \includegraphics[scale=0.7]{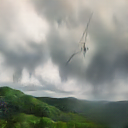}
    \end{subfigure}
    \begin{subfigure}[b]{0.2\textwidth}
        \centering
        \includegraphics[scale=0.7]{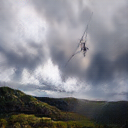}
    \end{subfigure}
    \vspace{0.1in}
    \begin{subfigure}[b]{0.2\textwidth}
        \centering
        \includegraphics[scale=0.7]{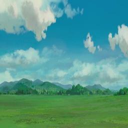}
    \end{subfigure}
    \begin{subfigure}[b]{0.2\textwidth}
        \centering
        \includegraphics[scale=0.7]{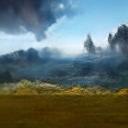}
    \end{subfigure}
    \begin{subfigure}[b]{0.2\textwidth}
        \centering
        \includegraphics[scale=0.7]{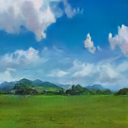}
    \end{subfigure}
    \begin{subfigure}[b]{0.2\textwidth}
        \centering
        \includegraphics[scale=0.7]{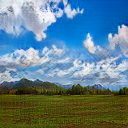}
    \end{subfigure}
    
    \vspace{0.1in}
    
    \begin{subfigure}[b]{0.2\textwidth}
        \centering
        \includegraphics[scale=0.7]{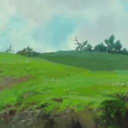}
    \end{subfigure}
    \begin{subfigure}[b]{0.2\textwidth}
        \centering
        \includegraphics[scale=0.7]{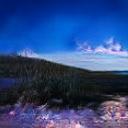}
    \end{subfigure}
    \begin{subfigure}[b]{0.2\textwidth}
        \centering
        \includegraphics[scale=0.7]{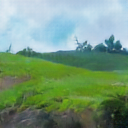}
    \end{subfigure}
    \begin{subfigure}[b]{0.2\textwidth}
        \centering
        \includegraphics[scale=0.7]{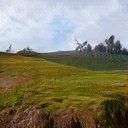}
    \end{subfigure}
    
    \vspace{0.1in}
    
    \begin{subfigure}[b]{0.2\textwidth}
        \centering
        \includegraphics[scale=0.7]{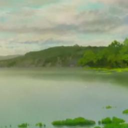}
        \caption{Input}
    \end{subfigure}
    \begin{subfigure}[b]{0.2\textwidth}
        \centering
        \includegraphics[scale=0.7]{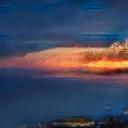}
        \caption{UNIT}
    \end{subfigure}
    \begin{subfigure}[b]{0.2\textwidth}
        \centering
        \includegraphics[scale=0.7]{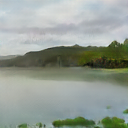}
        \caption{Cartoon-to-real}
    \end{subfigure}
    \begin{subfigure}[b]{0.2\textwidth}
        \centering
        \includegraphics[scale=0.7]{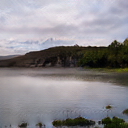}
        \caption{\textbf{toon2real}}
    \end{subfigure}
    \caption{Detailed comparisons in terms of contrast and content preservation. 
(a) Input images of cartoon scenes. 
(b) \textit{Result of UNIT}\cite{LiuBK17} which shows the lacking of content than (c) and (d). 
(c) \textit{Result of our previous work(Cartoon-to-real)}\cite{abs-1811-11796} which closely preserves the texture of input domain rather than translating into realistic one. 
(d) \textit{Result of our \textit{\textbf{toon2real}}} shows more contrast on content, compared to other works. The images were selected randomly.}
    \label{fig:result1}
\end{figure*}

\parindent 5ex \textbf{Dataset Collection:} Due to the lack of paired data between {cartoon} domain and {photo-realistic} domain, we took an approach to collect unpaired dataset for both domains. As deep learning is data hungry, initially, for \textit{realistic} domain, we scraped scenery images from \textit{Flickr} and many other sources which were tagged as \textit{scenery, sunrise, sunset, sea, sky} \& \textit{beach} and collected around $5000$ samples. Besides, for \textit{cartoon} domain, we extracted images from various Japanese anime movies and collected about $5000$ images. We extracted the scenery images from these movies consisting of \textit{sunsets, sea, sky, trees} etc. We excluded the frames which are darker to see, and the first and last few frames---as the introductory and credits part tend to be textual in a movie.
After hand-picking the appropriate images, in order to approximate with the size of the \textit{realistic} domain, we collected images from more than $15$ cartoon movies and clips, consisting of genres \textit{romance, spiritual, war, supernatural \& science-fiction}. For both the domain, images were of $128\times 128$ dimension. For the validation set, we took about $2500$ images from animation images and $2000$ from real-world photos. 

\parindent 5ex \textbf{Adversarial Loss:} Although in
\cite{GoodfellowPMXWOCB14}, a binary cross-entropy based Adversarial Loss function was proposed, we use a \textit{Least Squares Loss(LSGAN)} function for our training. According to \textit{Mao et. al} \cite{MaoLXLWS17}, we have explored that \textit{LSGAN} performs better in the case of vanishing gradient problem and thus shows more stability during training and produces much higher quality images in the case of \textit{Image-to-image Translation}. So, our adversarial loss stands as follows - 

\begin{equation}
For\, Generator\, G_r,\, \mathcal{L}_{G_r}\, =\, \frac{1}{m}\sum^m_{i=1}(1-D_r(G_r(c)))^2
\end{equation}

However, due to the deep similarities between cartoon and photo-realistic images, we observed that using only a single generator fails to map the differences between these $two$ domains. To resolve this issue, we use $two$ additional networks in our model, where a generator, $G_c$ tries to generate images of \textit{Cartoon} domain and a discriminator, $D_c$ tries to discriminate the generated image from \textit{cartoon} domain. The additional networks also perform according to the previously mentioned loss function.

\parindent 5ex \textbf{Reconstruction Loss:} Adding an additional generator solves the issue of mapping differences; however, it still lacks in content preservation of input domain. For this reason, we've used an additional loss function, by using the technique of forward and backward loss\cite{KimCKLK17, ZhuPIE17}. The motive of this function is that an image generated from an input can be reconstructed back to the input again such that $x = F(G(x))$, where $F$ and $G$ are generators and thus, it is able to map an image of target domain which is as close as possible to the image of input domain. In our paper, we call it \textit{Reconstruction Loss}. The equation is as follows - 
\begin{equation}
Forward\, Consistency\, Loss,\mathcal{L}_{f\_cyc} = \frac{1}{m} \sum^m_{i=1}(F_r(G_r(c)) - c)    
\end{equation}
\begin{equation}
Backward\, Consistency\, Loss,\mathcal{L}_{b\_cyc} = \frac{1}{m} \sum^m_{i=1}(G_r(G_c(r)) - r) 
\end{equation}

\parindent 5ex \textbf{Discriminator Normalization:}
Training GAN with efficiency is a hard nut to crack. Prior to previous works, it is known that discriminator tends to make the training slower and show more inconsistency during training. We used \textit{Spectral Normalization} technique, which was first proposed by \textit{Miyato et al.}\cite{abs-1802-05957}, to stabilize our training. The benefit of spectral normalization is that it doesn't need extra hyper-parameter tuning. Also, the computational cost is relatively small compared to other weight normalization techniques. \textit{Miyato et al.}\cite{abs-1802-05957} found better or same results with image generation tasks by utilizing this normalization technique. We can see from Figure~\ref{fig:fid_graph}  that, using this technique stabilized the training, where Figure~\ref{subfig:fid_spectral} is ours, which shows a much smoother curve of FID scores than Figure~\ref{subfig:fid_cycle}, which is the FID score-graph of our previous work(Cartoon-to-real)\cite{abs-1811-11796}. Also, we can see that ours achieved the least FID score within $145$ epoch, whereas our previous model\cite{abs-1811-11796} took more epochs for that.

\parindent 5ex \textbf{PatchGAN:}
As discriminators, we used PatchGAN which was first proposed in \textit{Isola et al.} \cite{cvpr}. The intuition of using this discriminator is that it works best for extracting the high-frequency details of the distribution. Another beneficial feature is, due to working on $N\times N$ patches, it takes fewer parameters and thus decreases the computation cost.

\section{Implementations \& Analysis} 
\label{sec:imp}
In this section, we discuss the implementation of our approach followed by illustrating its results.

\parindent 5ex \textbf{Network Structure:}
For generative networks we implemented the architecture from \textit{Johnson et al.} \cite{JohnsonAL16} who achieved amazing results for neural style transfer and super-resolution. The network includes two stride-2 convolutions, $6$ residual blocks\cite{HeZRS15}, and two fractional strided convolutions with stride $1/2$. We used instance normalization technique. For the discriminator network, we used $70\times70$ PatchGANs\cite{cvpr}. The network architecture of our proposed model is shown in Fig \ref{fig:cycle-gan-archi}.

\parindent 5ex \textbf{Evaluation Metric:}
We chose the Fr\'echet Inception Distance (FID) \cite{HeuselRUNKH17} for quantitative evaluation. As FID score measures the difference between the generated dataset and the target dataset, it has shown more consistency with human evaluation. Samples from \textit{P} and \textit{Q} are gone through an Inception-v3 network to transform it into feature space. Then, we calculate the Wasserstein-2 distance between the translated image and the real world images from an intermediate layer of an Inception-v3 network. The distance can be calculated as\\  
\begin{equation}
\centering
    FID = \left \| \mu_{x} - \mu_{y} \right \|_{2}^{2} + Tr( \Sigma_{x} + \Sigma_{y} - 2(\Sigma_{x}\Sigma_{y})^{\frac{1}{2}}) 
\end{equation}
where $(\mu_{x}, \Sigma_{x})$, and $(\mu_{x},\Sigma_{y})$ are the mean and covariance of the feature space of samples from \textit{P} and \textit{Q}.
Lower the FID score, the closer the distance between translated image and real domain images. As our task is image-to-image translation where we want our output to have the content of input cartoon images and the style of real-world images, we calculated a weighted average between them, where we used $80\%$ weight for target data and $20\%$ weight for input data. From Table \ref{tab:fid_table} we can see that our work has shown the least FID score compared to other state of the art model, i.e UNIT.

\begin{table}[h]
\centering
\resizebox{5cm}{!}{
\begin{tabular}{|l|l|}
\hline
\textit{Methods}   & \textit{FID Score}   \\ \hline
\textbf{Our work} & \textbf{39.2574} \\ \hline
Cartoon-to-real & 45.2566 \\ \hline
UNIT     & 55.9214 \\ \hline
\end{tabular}
}
\caption{FID score of our proposed model, in comparison with our previous work (Cartoon-to-real)\cite{abs-1811-11796} and UNIT model\cite{LiuBK17}.}
\label{tab:fid_table}
\end{table}

\begin{figure}[!htb]
\begin{center}
    \begin{subfigure}[b]{0.40\textwidth}
        \includegraphics[scale=0.38]{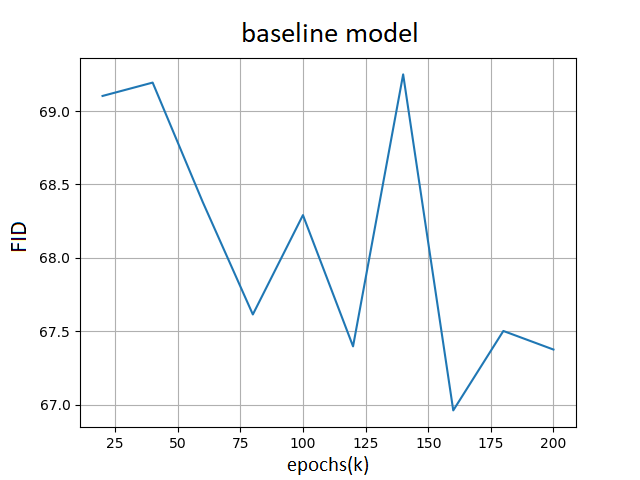}
        \caption{Cartoon-to-real}
        \label{subfig:fid_cycle}
    \end{subfigure}
    \begin{subfigure}[b]{0.40\textwidth}
        \includegraphics[scale=0.38]{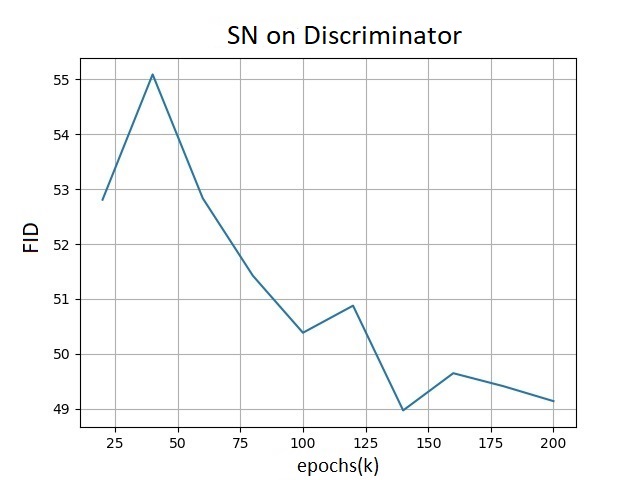}
        \caption{Our proposed method}
        \label{subfig:fid_spectral}
    \end{subfigure}
    \caption{Here, FID scores for our previous work (Cartoon-to-real) (a) and for our method (b) are shown from $20$ epochs up to $200$ epochs.}
    \label{fig:fid_graph}
\end{center}
\end{figure}

\parindent 5ex \textbf{Evaluation of Discriminator Normalization:}
By utilizing spectral normalization technique on discriminator network shown in Figure \ref{subfig:fid_spectral}, we started to gain a lower FID score from the very initial of training compared to our previous model. Spectral normalization is used on discriminator network which is shown in \ref{subfig:fid_spectral}. From \ref{subfig:fid_spectral}, the quality of transforming images doesn't improve monotonically during training. For example, the FID score of our work starts to drop at the $37$th epoch. On the contrary, previous model's FID score starts to rise after $125$th epoch and it crosses the initial FID scores, whereas, in our work, the scores didn't rise as the previous model did. From this, we can clarify that we achieved a more stabilized model and better scores. We can also clarify from Figure \ref{fig:fid_graph} that the stabilization technique also takes fewer training epochs to achieve better scores.

\parindent 5ex \textbf{Comparison with state of the art models:}
We compared our work with state of the art technique, i.e UNIT \cite{LiuBK17} and our previous work(Cartoon-to-real)\cite{abs-1811-11796}. In Figure \ref{fig:result1} we show a close-up view of an example, explaining that our work preserves much better vibrance and content preservation, whereas UNIT's\cite{LiuBK17} output shows a lacking of preserving content and information in image. Also our previous work\cite{abs-1811-11796} closely shows similar characteristics of input domain rather than realistic one.
\begin{figure}[!htb]
    \centering
    \begin{subfigure}[b]{0.2\textwidth}
        \centering
        \includegraphics[scale=0.7]{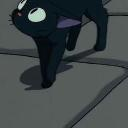}
        \caption{Cartoon scene}
    \end{subfigure}
    \begin{subfigure}[b]{0.2\textwidth}
        \centering
        \includegraphics[scale=0.7]{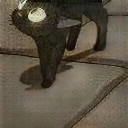}
        \caption{Our result}
    \end{subfigure}
    \caption{
    A failure case of our method. In the output image (b), the \textit{cat} remains cartoonish as in input (a) and is not translated into a realistic one.}
    \label{fig:limitation}
\end{figure}

\parindent 5ex \textbf{Limitations:} Despite achieving better FID score of all, it is still too high to be a perfect image translation score. In fact, we can see from Figure~\ref{fig:limitation} that, the output fails to achieve the
meaningful (semantically and geometrically)
structure of real-world objects---in this example a \textit{cat}.
This problem is also common in UNIT\cite{LiuBK17} and other image-to-image translation models.

\section{Conclusion}
In this paper, we showed a GAN based approach to translate images from cartoon domain to photo-realistic domain. We implemented our model based on CycleGAN\cite{ZhuPIE17}, where we used Reconstruction Loss for content preservation of input image and the PatchGAN for better texture extraction. By implementing spectral normalization technique on discriminator network, we showed that our model achieves better training stability and the lowest FID score of all the other models. Our future plan is to lessen our current limitations by investigating more geometry and content aware model to improve the texture so that the gap with the photo-realistic domain decreases. In addition to FID score, we have plans to arrange human-involved and perceptual evaluation processes to assess the correctness of our outcomes.

\bibliographystyle{IEEEtran}
\bibliography{main}

\begin{thebibliography}{10}
\providecommand{\url}[1]{#1}
\csname url@samestyle\endcsname
\providecommand{\newblock}{\relax}
\providecommand{\bibinfo}[2]{#2}
\providecommand{\BIBentrySTDinterwordspacing}{\spaceskip=0pt\relax}
\providecommand{\BIBentryALTinterwordstretchfactor}{4}
\providecommand{\BIBentryALTinterwordspacing}{\spaceskip=\fontdimen2\font plus
\BIBentryALTinterwordstretchfactor\fontdimen3\font minus
  \fontdimen4\font\relax}
\providecommand{\BIBforeignlanguage}[2]{{%
\expandafter\ifx\csname l@#1\endcsname\relax
\typeout{** WARNING: IEEEtran.bst: No hyphenation pattern has been}%
\typeout{** loaded for the language `#1'. Using the pattern for}%
\typeout{** the default language instead.}%
\else
\language=\csname l@#1\endcsname
\fi
#2}}
\providecommand{\BIBdecl}{\relax}
\BIBdecl

\bibitem{GoodfellowPMXWOCB14}
\BIBentryALTinterwordspacing
I.~J. Goodfellow, J.~Pouget{-}Abadie, M.~Mirza, B.~Xu, D.~Warde{-}Farley,
  S.~Ozair, A.~C. Courville, and Y.~Bengio, ``Generative adversarial nets,'' in
  \emph{Advances in Neural Information Processing Systems 27: Annual Conference
  on Neural Information Processing Systems 2014, December 8-13 2014, Montreal,
  Quebec, Canada}, 2014, pp. 2672--2680. [Online]. Available:
  \url{http://papers.nips.cc/paper/5423-generative-adversarial-nets}
\BIBentrySTDinterwordspacing

\bibitem{ZhuPIE17}
\BIBentryALTinterwordspacing
J.~Zhu, T.~Park, P.~Isola, and A.~A. Efros, ``Unpaired image-to-image
  translation using cycle-consistent adversarial networks,'' in \emph{{IEEE}
  International Conference on Computer Vision, {ICCV} 2017, Venice, Italy,
  October 22-29, 2017}, 2017, pp. 2242--2251. [Online]. Available:
  \url{https://doi.org/10.1109/ICCV.2017.244}
\BIBentrySTDinterwordspacing

\bibitem{KimCKLK17}
\BIBentryALTinterwordspacing
T.~Kim, M.~Cha, H.~Kim, J.~K. Lee, and J.~Kim, ``Learning to discover
  cross-domain relations with generative adversarial networks,'' in
  \emph{Proceedings of the 34th International Conference on Machine Learning,
  {ICML} 2017, Sydney, NSW, Australia, 6-11 August 2017}, 2017, pp. 1857--1865.
  [Online]. Available: \url{http://proceedings.mlr.press/v70/kim17a.html}
\BIBentrySTDinterwordspacing

\bibitem{abs-1802-05957}
\BIBentryALTinterwordspacing
T.~Miyato, T.~Kataoka, M.~Koyama, and Y.~Yoshida, ``Spectral normalization for
  generative adversarial networks,'' \emph{CoRR}, vol. abs/1802.05957, 2018.
  [Online]. Available: \url{http://arxiv.org/abs/1802.05957}
\BIBentrySTDinterwordspacing

\bibitem{JohnsonAL16}
\BIBentryALTinterwordspacing
J.~Johnson, A.~Alahi, and F.~Li, ``Perceptual losses for real-time style
  transfer and super-resolution,'' \emph{CoRR}, vol. abs/1603.08155, 2016.
  [Online]. Available: \url{http://arxiv.org/abs/1603.08155}
\BIBentrySTDinterwordspacing

\bibitem{abs-1810-04991}
\BIBentryALTinterwordspacing
X.~Yu, X.~Cai, Z.~Ying, T.~H. Li, and G.~Li, ``Singlegan: Image-to-image
  translation by a single-generator network using multiple generative
  adversarial learning,'' \emph{CoRR}, vol. abs/1810.04991, 2018. [Online].
  Available: \url{http://arxiv.org/abs/1810.04991}
\BIBentrySTDinterwordspacing

\bibitem{LiuBK17}
\BIBentryALTinterwordspacing
M.~Liu, T.~Breuel, and J.~Kautz, ``Unsupervised image-to-image translation
  networks,'' \emph{CoRR}, vol. abs/1703.00848, 2017. [Online]. Available:
  \url{http://arxiv.org/abs/1703.00848}
\BIBentrySTDinterwordspacing

\bibitem{ZhangXLZHWM16}
\BIBentryALTinterwordspacing
H.~Zhang, T.~Xu, H.~Li, S.~Zhang, X.~Huang, X.~Wang, and D.~N. Metaxas,
  ``Stackgan: Text to photo-realistic image synthesis with stacked generative
  adversarial networks,'' \emph{CoRR}, vol. abs/1612.03242, 2016. [Online].
  Available: \url{http://arxiv.org/abs/1612.03242}
\BIBentrySTDinterwordspacing

\bibitem{ReedAYLSL16}
\BIBentryALTinterwordspacing
S.~E. Reed, Z.~Akata, X.~Yan, L.~Logeswaran, B.~Schiele, and H.~Lee,
  ``Generative adversarial text to image synthesis,'' \emph{CoRR}, vol.
  abs/1605.05396, 2016. [Online]. Available:
  \url{http://arxiv.org/abs/1605.05396}
\BIBentrySTDinterwordspacing

\bibitem{yi2017dualgan}
Z.~Yi, H.~R. Zhang, P.~Tan, and M.~Gong, ``Dualgan: Unsupervised dual learning
  for image-to-image translation.'' in \emph{ICCV}, 2017, pp. 2868--2876.

\bibitem{he2016dual}
D.~He, Y.~Xia, T.~Qin, L.~Wang, N.~Yu, T.-Y. Liu, and W.-Y. Ma, ``Dual learning
  for machine translation,'' in \emph{Advances in Neural Information Processing
  Systems}, 2016, pp. 820--828.

\bibitem{liu2016coupled}
M.-Y. Liu and O.~Tuzel, ``Coupled generative adversarial networks,'' in
  \emph{Advances in neural information processing systems}, 2016, pp. 469--477.

\bibitem{kingma2013auto}
D.~P. Kingma and M.~Welling, ``Auto-encoding variational bayes,'' \emph{arXiv
  preprint arXiv:1312.6114}, 2013.

\bibitem{larsen2015autoencoding}
A.~B.~L. Larsen, S.~K. S{\o}nderby, H.~Larochelle, and O.~Winther,
  ``Autoencoding beyond pixels using a learned similarity metric,'' \emph{arXiv
  preprint arXiv:1512.09300}, 2015.

\bibitem{rezende2014stochastic}
D.~J. Rezende, S.~Mohamed, and D.~Wierstra, ``Stochastic backpropagation and
  variational inference in deep latent gaussian models,'' in
  \emph{International Conference on Machine Learning}, vol.~2, 2014.

\bibitem{GulrajaniAADC17}
\BIBentryALTinterwordspacing
I.~Gulrajani, F.~Ahmed, M.~Arjovsky, V.~Dumoulin, and A.~C. Courville,
  ``Improved training of wasserstein gans,'' \emph{CoRR}, vol. abs/1704.00028,
  2017. [Online]. Available: \url{http://arxiv.org/abs/1704.00028}
\BIBentrySTDinterwordspacing

\bibitem{RadfordMC15}
\BIBentryALTinterwordspacing
A.~Radford, L.~Metz, and S.~Chintala, ``Unsupervised representation learning
  with deep convolutional generative adversarial networks,'' \emph{CoRR}, vol.
  abs/1511.06434, 2015. [Online]. Available:
  \url{http://arxiv.org/abs/1511.06434}
\BIBentrySTDinterwordspacing

\bibitem{karras2017progressive}
T.~Karras, T.~Aila, S.~Laine, and J.~Lehtinen, ``Progressive growing of gans
  for improved quality, stability, and variation,'' \emph{arXiv preprint
  arXiv:1710.10196}, 2017.

\bibitem{Arjovsky2017WassersteinG}
M.~Arjovsky, S.~Chintala, and L.~Bottou, ``Wasserstein gan,'' \emph{CoRR}, vol.
  abs/1701.07875, 2017.

\bibitem{metz2016unrolled}
L.~Metz, B.~Poole, D.~Pfau, and J.~Sohl-Dickstein, ``Unrolled generative
  adversarial networks,'' \emph{arXiv preprint arXiv:1611.02163}, 2016.

\bibitem{salimans2016weight}
T.~Salimans and D.~P. Kingma, ``Weight normalization: A simple
  reparameterization to accelerate training of deep neural networks,'' in
  \emph{Advances in neural information processing systems}, 2016, pp. 901--909.

\bibitem{abs-1811-11796}
\BIBentryALTinterwordspacing
K.~M.~A. Sultan, L.~K. Rupty, N.~I. Pranto, S.~K. Shuvo, and M.~I. Jubair,
  ``Cartoon-to-real: An approach to translate cartoon to realistic images using
  {GAN},'' \emph{CoRR}, vol. abs/1811.11796, 2018. [Online]. Available:
  \url{http://arxiv.org/abs/1811.11796}
\BIBentrySTDinterwordspacing

\bibitem{MaoLXLWS17}
\BIBentryALTinterwordspacing
X.~Mao, Q.~Li, H.~Xie, R.~Y.~K. Lau, Z.~Wang, and S.~P. Smolley, ``Least
  squares generative adversarial networks,'' in \emph{{IEEE} International
  Conference on Computer Vision, {ICCV} 2017, Venice, Italy, October 22-29,
  2017}, 2017, pp. 2813--2821. [Online]. Available:
  \url{https://doi.org/10.1109/ICCV.2017.304}
\BIBentrySTDinterwordspacing

\bibitem{cvpr}
\BIBentryALTinterwordspacing
\emph{2017 {IEEE} Conference on Computer Vision and Pattern Recognition, {CVPR}
  2017, Honolulu, HI, USA, July 21-26, 2017}.\hskip 1em plus 0.5em minus
  0.4em\relax {IEEE} Computer Society, 2017. [Online]. Available:
  \url{http://ieeexplore.ieee.org/xpl/mostRecentIssue.jsp?punumber=8097368}
\BIBentrySTDinterwordspacing

\bibitem{HeZRS15}
\BIBentryALTinterwordspacing
K.~He, X.~Zhang, S.~Ren, and J.~Sun, ``Deep residual learning for image
  recognition,'' \emph{CoRR}, vol. abs/1512.03385, 2015. [Online]. Available:
  \url{http://arxiv.org/abs/1512.03385}
\BIBentrySTDinterwordspacing

\bibitem{HeuselRUNKH17}
\BIBentryALTinterwordspacing
M.~Heusel, H.~Ramsauer, T.~Unterthiner, B.~Nessler, G.~Klambauer, and
  S.~Hochreiter, ``Gans trained by a two time-scale update rule converge to a
  nash equilibrium,'' \emph{CoRR}, vol. abs/1706.08500, 2017. [Online].
  Available: \url{http://arxiv.org/abs/1706.08500}
\BIBentrySTDinterwordspacing

\end{thebibliography}

\end{document}